\documentclass[10pt,twocolumn,letterpaper]{article}

\usepackage{cvpr}      % To produce the REVIEW version
\usepackage{graphicx}
\usepackage{amsmath}
\usepackage{booktabs}
\usepackage{amsfonts}
\usepackage{multirow}
\usepackage{utfsym}
\usepackage{ulem}
\usepackage{float}

\usepackage[pagebackref,breaklinks,colorlinks]{hyperref}

\usepackage[capitalize]{cleveref}
\crefname{section}{Sec.}{Secs.}
\Crefname{section}{Section}{Sections}
\Crefname{table}{Table}{Tables}
\crefname{table}{Tab.}{Tabs.}

\def\cvprPaperID{*****} % *** Enter the CVPR Paper ID here
\def\confName{CVPR}
\def\confYear{2022}

\begin{document}

\title{The Championship-Winning Solution for the 5th CLVISION Challenge 2024}

\author{
Sishun Pan,
Tingmin Li,
Yang Yang$^*$
\\Nanjing University of Science and Technology
}

\maketitle
%%%%%%%%% ABSTRACT
\begin{abstract}
    In this paper, we introduce our approach to the 5th CLVision Challenge, which presents distinctive challenges beyond traditional class incremental learning. Unlike standard settings, this competition features the recurrence of previously encountered classes and includes unlabeled data that may contain Out-of-Distribution (OOD) categories. Our approach is based on Winning Subnetworks to allocate independent parameter spaces for each task addressing the catastrophic forgetting problem in class incremental learning and employ three training strategies: supervised classification learning, unsupervised contrastive learning, and pseudo-label classification learning to fully utilize the information in both labeled and unlabeled data, enhancing the classification performance of each subnetwork. Furthermore, during the inference stage, we have devised an interaction strategy between subnetworks, where the prediction for a specific class of a particular sample is the average logits across different subnetworks corresponding to that class, leveraging the knowledge learned from different subnetworks on recurring classes to improve classification accuracy. These strategies can be simultaneously applied to the three scenarios of the competition, effectively solving the difficulties in the competition scenarios. Experimentally, our method ranks first in both the pre-selection and final evaluation stages, with an average accuracy of 0.4535 during the preselection stage and an average accuracy of 0.4805 during the final evaluation stage.
\end{abstract}

%%%%%%%%% BODY TEXT
\section{Introduction}
In traditional class-Incremental settings\cite{ccl_survey}, each task contains a labeled dataset, with non-intersecting class sets between tasks. Models must continuously adapt to new classes while retaining knowledge of previous ones. In competitive scenarios, training data for each task comprises both labeled and unlabeled samples. This data includes classes from both new and previously encountered tasks. Moreover, the unlabeled data may contain categories that are out-of-distribution (OOD). Yang et al.\cite{CILF} address the issues of out-of-distribution detection and model expansion in Class-Incremental Learning (CIL) by proposing a Class-Incremental Learning without Forgetting (CILF) framework, which aims to learn adaptive embeddings for processing novel class detection and model updates within a unified framework.

To address the problem of open-set semi-supervised class-incremental learning with repetition, we decompose the problem into four sub-problems: conducting traditional class-incremental learning, utilizing labeled data, utilizing unlabeled data, and handling repeated classes. (1) For class-incremental learning, we employ the parameter isolation method Winning Subnetworks\cite{wsn}, allowing each task to learn within its parameter space, effectively mitigating catastrophic forgetting. (2) To utilize labeled data, we train labeled data using cross-entropy loss. (3) Regarding the utilization of unlabeled data, we adopt two strategies. Firstly, employing unsupervised contrastive learning to extract discriminative and robust features from unlabeled data. Secondly, leveraging fixmatch\cite{fixmatch} and high threshold constraint to identify confident unlabeled data, utilizing their pseudo-labels for enhanced classification performance. (4) For handling repeated classes, we employ an subnetworks interaction strategy. We compute the average logits across multiple sub-networks to obtain more robust predictions for samples within specific classes. As a result, our method ranks first in both the pre-selection and final evaluation stages, producing an average accuracy of 0.4535 in the pre-selection phase and an average accuracy of 0.4805 in the final evaluation phase. In the remainder of this technical report, we will introduce the detailed architecture of our solution for this challenge.
\label{sec:intro}

\begin{figure*}[!h]
	\centering
	\includegraphics[width=\linewidth]{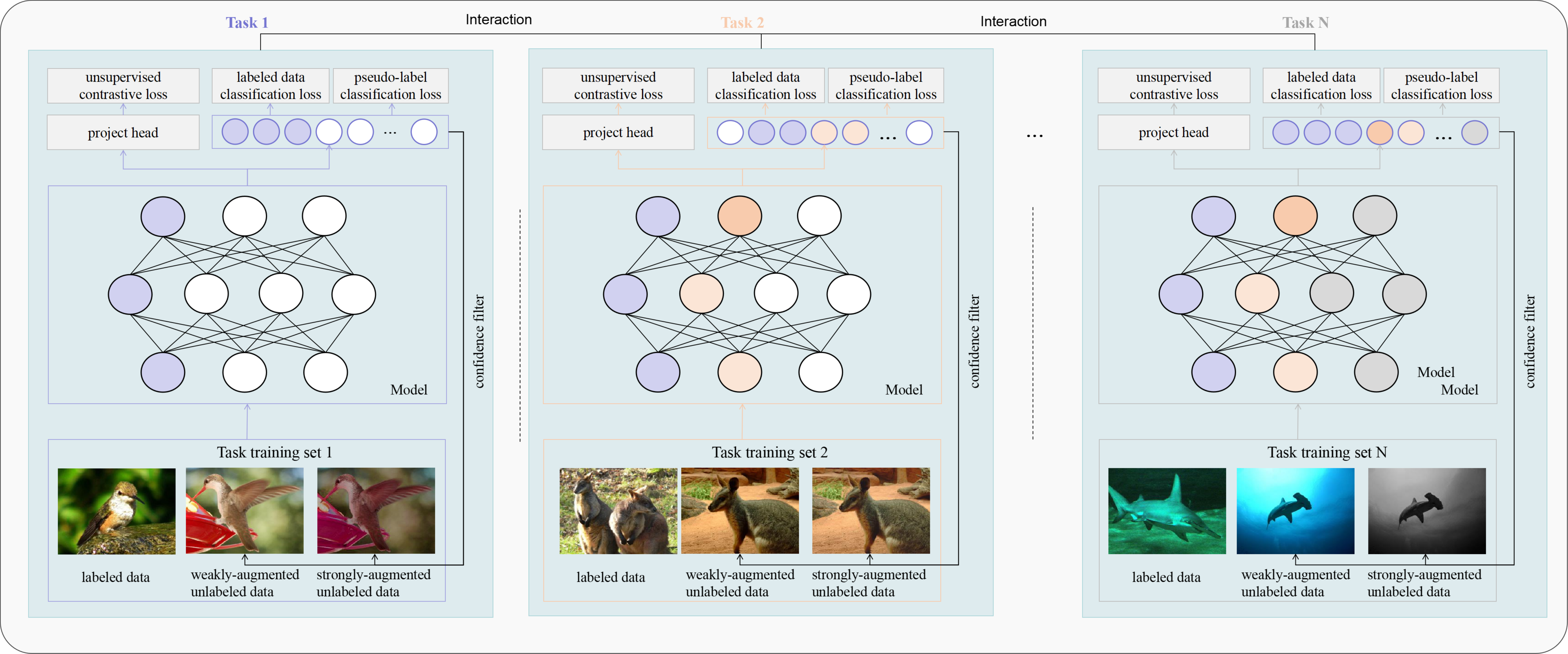}
	\caption{Overall Architecture. Our solution consists of three core components, which includes WSN-based Subnetwork Partitioning, three training strategies and subnetworks interaction.}\label{fig:overview}
\end{figure*}

%------------------------------------------------------------------------
\section{Related Work}
\subsection{Continual learning}
Continual learning\cite{wang2024comprehensive} refers to a model continuously learning a series of tasks while retaining knowledge from previous tasks. There are five main approaches to continual learning: Regularization-based methods\cite{ewc}\cite{afec}\cite{connector} add explicit regularization terms to balance the knowledge between new and old tasks. Replay-based\cite{replay_1}\cite{replay_2} methods store samples from old tasks in memory or use additional generative models to replay historical data. Optimization-based methods\cite{GEM}\cite{OGD}\cite{Adam-NSCL} use specially designed optimization processes to reduce forgetting, such as ensuring that the gradient directions of model updates are orthogonal to the feature space of historical task data. Representation-based methods\cite{CaSSLe}\cite{Co2L} aim to prevent catastrophic forgetting by leveraging the advantages of learned representations. Architecture-based methods\cite{hat}\cite{yang2021cost}\cite{yang2019adaptive}\cite{wsn} are divided into two main forms: parameter isolation and dynamic architecture. Both approaches explicitly address catastrophic forgetting by constructing task-specific parameters.

\subsection{Semi-supervised learning}
Semi-supervised learning aims to train models using a large amount of unlabeled data\cite{he2022not}. Consistency regularization and pseudo-labeling techniques are widely used in SSL algorithms. Pan et al.\cite{wad} propose WAD method, which captures adaptive weights and high-quality pseudo labels to target instances by exploring point mutual information (PMI) in representation space to maximize the role of unlabeled data and filter unknown categories. Yang et al.\cite{yang2021s2osc} develop a novel OSC algorithm, S2OSC, which trains a holistic classification model by combing in-class and out-of-class labeled data with the remaining unlabeled test data in a semi-supervised paradigm.

\subsection{OOD detection}
OOD detection aims to enable the model to identify test samples that deviate significantly from the training distribution. Designing a score function is the most commonly used method in OOD detection tasks. Yang et al.\cite{wiseopen} propose Wise Open-set Semi-supervised Learning (WiseOpen), a generic OSSL framework that selectively leverages the open-set data for training the model. Ming et al.\cite{ming2022poem} propose propose a novel posterior sampling-based outlier mining framework, POEM, which facilitates efficient use of outlier data and promotes learning a compact decision boundary between ID and OOD data for improved detection. Xi et al.\cite{xi2023robust} propose an open-world SSL method for Self-learning Open-world Classes (SSOC), which can explicitly self-learn multiple unknown classes.
\label{sec:Related}

%-------------------------------------------------------------------------

%-------------------------------------------------------------------------

%------------------------------------------------------------------------
\section{Methodology}
Figure \ref{fig:overview} illustrates the overall architecture of our solution, which comprises three core components: (1) structural Design, a subnetwork partitioning method based on the WSN\cite{wsn}; (2) three training strategies, including labeled data classification learning, unsupervised contrastive learning, and pseudo-label classification learning; (3) subnetworks interaction for the inference stage. Next, we will specifically introduce the above three components, as well as how to integrate these strategies into the ResNet-18 Backbone. 

\subsection{WSN-based Subnetwork Partitioning}
We choose the WSN method\cite{wsn}, a parameter isolation method, as the foundational approach to perform continual learning. We select it for two reasons: Firstly, as a parameter isolation method, WSN fundamentally eliminates catastrophic forgetting by assigning dedicated parameters to each task. Secondly, WSN can selectively reuse parameters from previous tasks, effectively leveraging knowledge of repeated classes across different tasks and reducing training convergence time. Furthermore, we found that the WSN method still has shortcomings. For example, the parameter importance matrix of the current task is initialized with the parameter importance matrix of the previous task, thereby it tends to use the parameters used in previous task, while ignoring the use of other paramters. Therefore, we have adopted a gradient supplementation strategy. Specifically, in the learning of WSN, each weight \(\theta\) associated with an importance score \(s\), where the update process of \(s\) is as follows: 
\[
\mathbf{s} \leftarrow \mathbf{s} - \begin{cases}
\eta \left( \frac{\partial \mathcal{L}}{\partial \mathbf{s}} \right) \cdot \gamma & \text{if } \mathbf{M}_{t-1} = 1 \text{ or } \mathbf{m}_t = 0 \\
\eta \left( \frac{\partial \mathcal{L}}{\partial \mathbf{s}} \right) & \text{otherwise}
\end{cases}
\]
, where \(\eta\) is the learning rate, \(\gamma\) is the gradient supplementation coefficient, \(\mathbf{M}_{t-1}\) is the historical task mask, and \(\mathbf{m}_t\) is the current task mask. In this way, the gradient of parameters not selected for the current task is supplemented, which is more beneficial for learning the current task.

\subsection{Loss function design}
\textbf{labeled data classification learning} aims to fully leverage the information contained in labeled dataset to train individual subnetworks for each task. \begin{figure}[t]
	\centering
	\includegraphics[width=\linewidth]{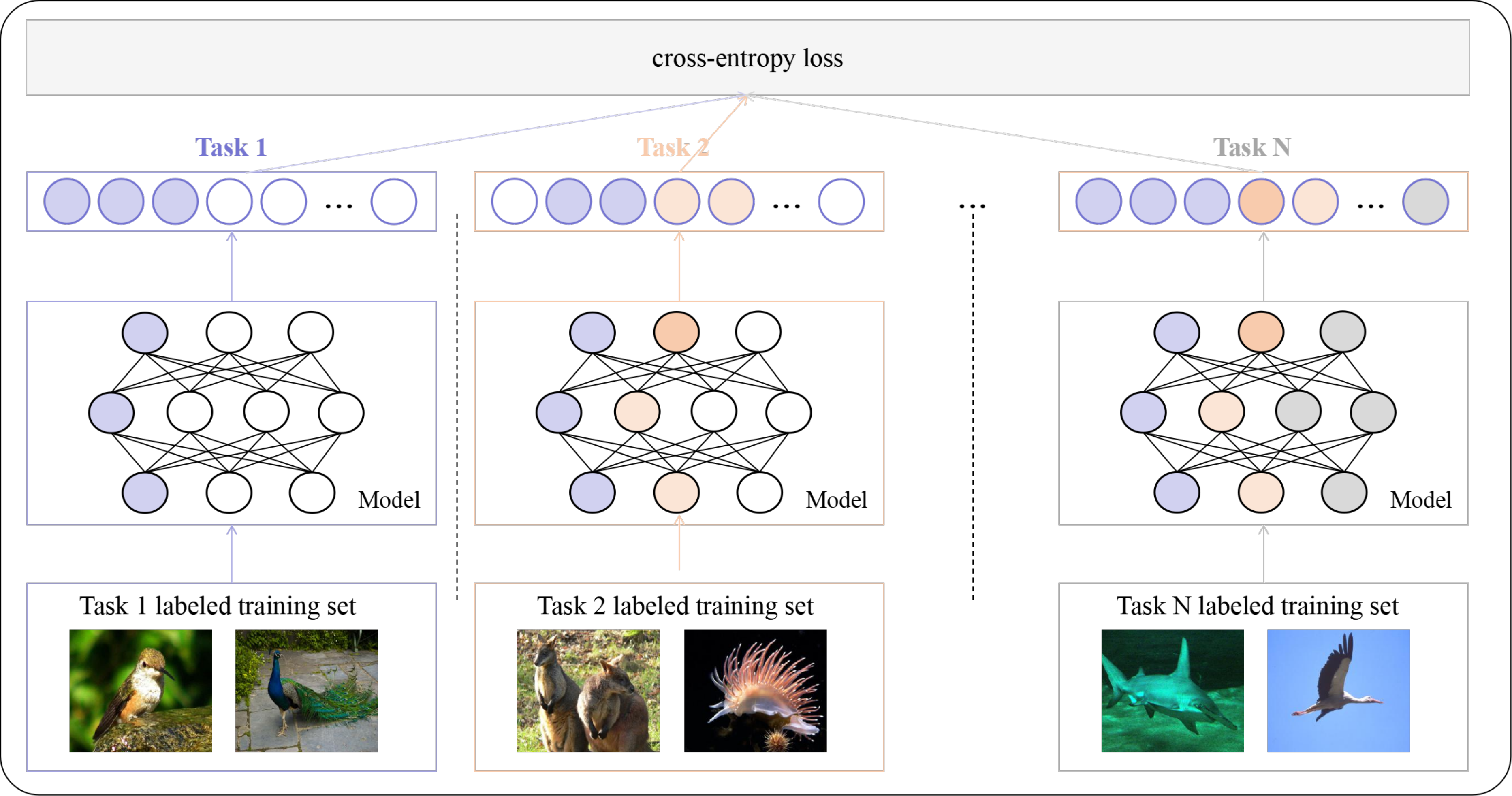}
	\caption{The process of labeled data classification learning.}\label{fig:label_loss}
\end{figure}As shown in Figure \ref{fig:label_loss}, by computing the cross-entropy loss between the logits predicted by the subnetwork and the actual labels in the labeled dataset, we optimize the classification performance of the subnetwork. The loss function is defined as follows:
\[L_l = -\frac{1}{N_{l}^t} \sum_{i=1}^{N_{l}^t} y_{i, l}^t \log(f(x_{i, l}^t; \theta))\] 
where \(N_{l}^t\) represents the number of labeled data for task \(t\), \(y_{i, l}^t\) corresponds to the label of the \(i\)-th labeled data for task \(t\), and \(\mathbf{x}_{i, l}^t\) represents the image corresponding to the \(i\)-th labeled data for task \(t\), and \(f(\cdot; \theta)\) denotes the neural network parameterized by the model weights \(\theta\).

\begin{figure}[t]
	\centering
	\includegraphics[width=\linewidth]{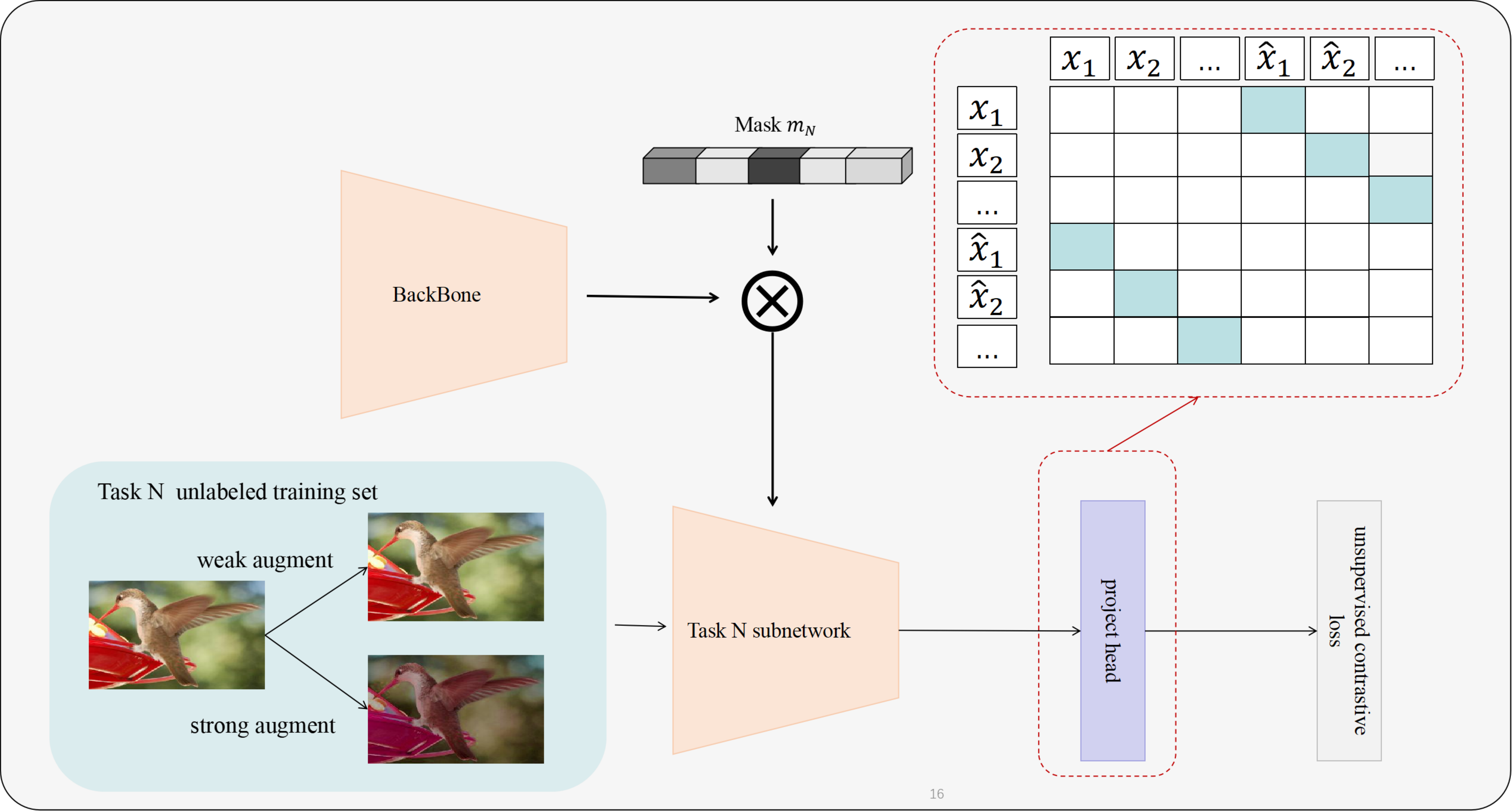}
	\caption{The process of unsupervised contrastive learning.}\label{fig:unlabel_loss}
\end{figure}

\textbf{Unsupervised contrastive learning} aims to leverage unlabeled data to learn more robust features, addressing the challenge of insufficient labeled data for learning effective feature representations. Specifically, we initially construct both weak augmentation and strong augmentation forms for each unlabeled data, thereby a batch of unlabeled data can be represented as 
\(\hat{\mathcal{B}} = \{ x_1, \hat{x}_1, x_2, \hat{x}_2, \dots, x_N, \hat{x}_N \}\), where $N$ represent the batchsize of unlabeled dataset, \(x_i\) and \(\hat{x_i}\) respectively represent the weakly augmented and strongly augmented versions of the unlabeled data. We adopt the following loss function:
\begin{align*}
L_u =\sum_{i=1}^{N} -\Biggl(& \log \left( \frac{\exp^{S(x_i, \hat{x}_i)}}{\sum_{z \in \hat{\mathcal{B}} \setminus \{x_i\}} \exp^{S(x, z)} + \epsilon} \right) \\
&+ \log \left( \frac{\exp^{S(x_i, \hat{x}_i)}}{\sum_{z \in \hat{\mathcal{B}} \setminus \{\hat{x}_i\}} \exp^{S(z, \hat{x}_i)} + \epsilon} \right) + 2 \cdot \epsilon \Biggr)
\end{align*}, where \(S(x, y)\) represents the dot product similarity between the logits obtained from input \(x\) and logits obtained from input \(y\). The optimization objective of this loss function involves two aspects: Firstly, to minimize the logits difference between the weakly-augmented and strongly-augmented version of the same unlabeled data. Secondly, we aim to maximize the logits difference between the original image and other samples within the batch, as well as between the augmented image and other samples within the batch. This process is shown in the figure \ref{fig:unlabel_loss}, 
maintaining consistency in inherent features during the transformation process and enhancing the model's robustness and generalization capability.

\begin{figure*}[!h]
	\centering
	\includegraphics[width=\linewidth]{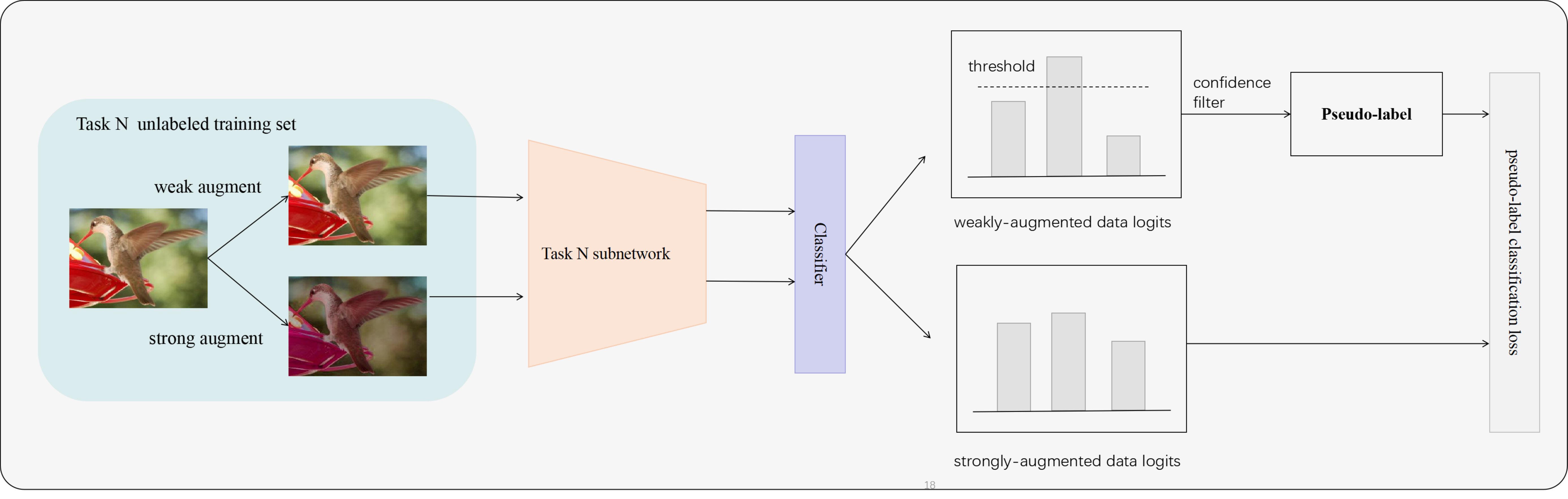}
	\caption{The process of pseudo-label classification learning.}\label{fig:p_loss}
\end{figure*}
\textbf{Pseudo-label classification learning} aims to leverage high-confidence pseudo-label of unlabeled data to enhance classification performance. As shown in figure \ref{fig:p_loss}, we first calculate the class distribution of strongly enhanced and weakly enhanced versions of unlabeled data, represented as $f(x; \theta)$, $f(\hat{x}; \theta)$ respectively, and use $\hat{q_b} = argmax(f(x; \theta))$ as the pseudo-label, we enforce the cross-entropy loss on the model's output for strongly-augmented version of the unlabeled data and pseudo-label. Meanwhile, in order to filter out low-confidence unlabeled data, we set a threshold. Only when the maximum probability value in the output exceeds the threshold can the data be considered to be of high confidence. This can be formalized as:
\[
L_p = \frac{1}{N} \sum_{i=1}^{ N} \mathbb{I}{(\max(q_b) > \epsilon)} H(\hat{q_b}, f(\hat{x}; \theta)))
\]
where \(N\) represents the batchsize of unlabeled dataset, $H$ is cross-entropy loss, and $\epsilon$ represents a threshold used to filter unlabeled data with low confidence. This optimization objectives aims to minimize the cross-entropy loss between the class distribution of strongly augmented versions and their pseudo-labels of high-confidence unlabeled samples, effectively enhancing the utilization of unlabeled data and demonstrates that OOD data contained within unlabeled data do not always impair classification performance\cite{yang2024not}. Subsequently, the final loss function can be expressed as follows: \[
\mathcal{L} = L_l + \alpha \cdot L_u + \beta \cdot L_p
\], where $\alpha$ and $\beta$ are hyperparameters.

\subsection{Interaction between subnetworks}
\begin{figure}[t]
	\centering
	\includegraphics[width=\linewidth]{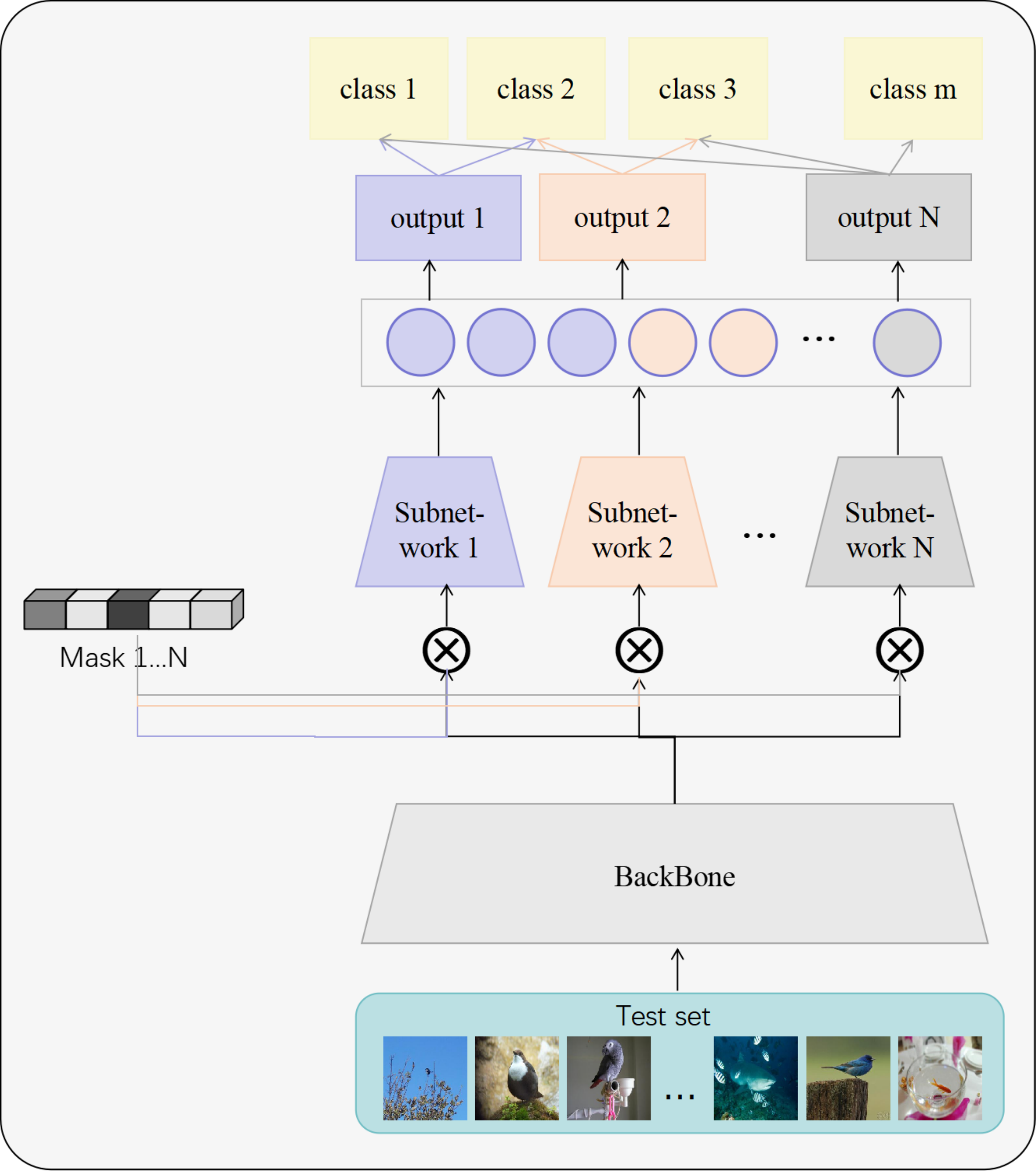}
	\caption{The process of subnetworks interaction.}\label{fig:ensemble}
\end{figure}
Interaction between subnetworks is the last stage, shown in the figure \ref{fig:ensemble}. Due to the occurrence of repeated categories in different experience under the competition setting, during the inference stage, to leverage the diverse knowledge within all experiences, we will calculate the logits of the sample output across all subnetworks, then place the logits output by each subnetwork into the stack of each respective class. The final logits of a specific class of the sample is the average of all logits in the stack for that class. 

\section{Experiments}
\subsection{Implementation Detail}
\textbf{Training Setting.} We employ ResNet-18 as the backbone with input images resized to \textit{224$\times$224}. We use an AdamW optimizer with a learning rate of 3e-4 and weight decay of 1e-4. We decay the learning rate with Cosine Annealing Learning Rate Scheduler, which dynamically adjusts the learning rate based on the cosine annealing schedule. The batch size for labeled data is set to 32, while the batch size for unlabeled data is set to 64. Training duration varies: 200 epochs for the first task, 100 epochs for tasks 2 through 5, and 50 epochs for subsequent tasks.

\textbf{Test Setting.} For testing, the batch size is set to 256. We conduct inference on samples across each sub-network and then combine the average predictions of each class from different sub-networks to obtain the final prediction for each class.

\subsection{Result}

\begin{table}[!ht]
\centering
\caption{We report the average accuracy of our methods on the pre-selection phase.}
\begin{tabular}{ccc}% 其中，tabular是表格内容的环境；c表示centering，即文本格式居中；c的个数代表列的个数
\toprule[1.5pt]
\# & \textbf{Method}  &  \textbf{Average Accuracy} \\ %换行
\midrule %[2pt]  
1 & Baseline & 35+ \\
2 & Unsupervised contrastive learning & 40+ \\
3 & Pseudo-label classification learning & 42+ \\
4 & Subnetworks interaction & 45+ \\
\bottomrule[1.5pt]
\label{table-1}
\end{tabular}
\end{table}
With Original WSN
As reported in table \ref{table-1}, with WSN base method and labeled data classification learning, the average accuracy reached 35+ score. The most significant improvement strategy is the unsupervised contrastive learning, which improves the accuracy by 5+ points. By utilizing pseudo-label classification learning and subnetworks interaction, we achieved the highest average accuracy score on the leaderboard of 45+.

\section{Conclusion} 
This report summarizes our solution for 5th CLVISION Challenge, which includes three core components: WSN-based Subnetwork Partitioning, three major training strategies (labeled data classification learning, unsupervised contrastive learning, and pseudo-label classification learning), and subnetworks interaction. Our solution overcomes the catastrophic forgetting problem through the parameter isolation method Winning Subnetworks and effectively utilizes unlabeled data for enhancing the learning ability of model, balancing model plasticity and stability. The final competition results show the effectiveness of our solution.
%%%%%%%%% REFERENCES
{\small
\bibliographystyle{ieee_fullname}
\bibliography{main}
}

\end{document}